\documentclass[manuscript,screen]{acmart}
\usepackage{graphicx}
\newcommand{\ours}{ZeroVLM}
\newcommand{\Mini}{MiniGPT-4}
\newcommand{\llava}{LLaVA}
\newcommand{\llama}{LLaMA}
\newcommand{\vsr}{VSR}
\newcommand{\WU}{What'sUp}
\newcommand{\IGPT}{InternGPT}
\newcommand{\GPT}{GPT-4}
\newcommand{\vqa}{VQA}
\newcommand{\json}{JSON}
\newcommand{\VLMs}{VLMs}
\newcommand{\VLM}{VLM}
\usepackage{booktabs}
\usepackage{array}

\usepackage{multirow}
\usepackage{caption}
\usepackage{makecell}

\usepackage{tikz}
\usepackage{pgfplots}
\usepackage{caption}
\usepackage{subcaption}

\usepackage{amsmath}
\usepackage{natbib}
\usepackage{graphicx}

\usepackage{xcolor}
\usepackage{hyperref}

\definecolor{customblue}{RGB}{111,124,167}
\definecolor{customorange}{RGB}{196,146,123}
\AtBeginDocument{%
  }

\acmISBN{978-1-4503-XXXX-X/18/06}

\begin{document}

\title{I Know About “Up”! Enhancing Spatial Reasoning in Visual Language Models Through 3D Reconstruction}

\author{Hao Zhou}
\email{hao.zhou28@outlook.com}
\affiliation{%
  \institution{Guangdong Polytechnic Normal University}
  \country{China}
}

\author{Zaiqiao Meng}
\authornote{Corresponding Author.}
\affiliation{%
  \institution{University of Glasgow }
  \country{UK}}
\email{zaiqiao.meng@glasgow.ac.uk}

\author{Yifang Chen}
\affiliation{%
  \institution{Guangdong Polytechnic Normal University}
  \country{China}
}
\email{chenyf@gpnu.edu.cn}



\begin{abstract}
  Visual Language Models (\VLMs{}) are essential for various tasks, particularly the visual reasoning tasks, due to their robust multi-modal information integration, visual reasoning capabilities, and contextual awareness. However, existing \VLMs{}' visual spatial reasoning capabilities are often inadequate, struggling even with basic tasks such as distinguishing left from right. To address this, we propose the \ours{}\footnote{Code: \url{https://github.com/zhouhao028/Iknow_up}} model, designed to enhance the visual spatial reasoning abilities of \VLMs{}. \ours{} employs Zero-1-to-3, a 3D reconstruction model for obtaining different views of the input images and incorporates a view prompt to further improve visual spatial reasoning. Experimental results on four visual spatial reasoning datasets show that our \ours{} achieves up to 19.48\% accuracy improvement, which indicates the effectiveness of 3D reconstruction and view prompt of our \ours{}.
\end{abstract}

\keywords{Visual Spatial 
	Reasoning, 3D 
	Reconstruction, Visual
	Language Models}

\maketitle

\section{Introduction}
Visual Language Models (\VLMs{}), such as \llava{}~\citep{liu2024visual}, \Mini{}~\citep{zhu2023minigpt} and \IGPT{}~\citep{liu2023internchat}, are a class of deep neural models adept at simultaneously processing and understanding both visual and linguistic information. These \VLMs{} often consist of an image encoder, an embedding projector to align image and text representations and a text decoder to process the projected image embedding and text representations, enabling joint understanding and reasoning between these modalities~\citep{li2022vision}. 
By using the language generation power of underlining text decoders, these \VLMs{} showcase remarkable interaction capabilities in various applications, such as referring expression comprehension~\citep{li2019visualbert,yu2016modeling}, visual question answering~\citep{Goyal2017,zellers2019recognition,hudson2019gqa}, visual language reasoning~\citep{liu2022deplot}, and entailment~\citep{liu2021visually,xie2019visual}.


\begin{figure}[t]
	\centering
	\includegraphics[width=0.5\columnwidth]{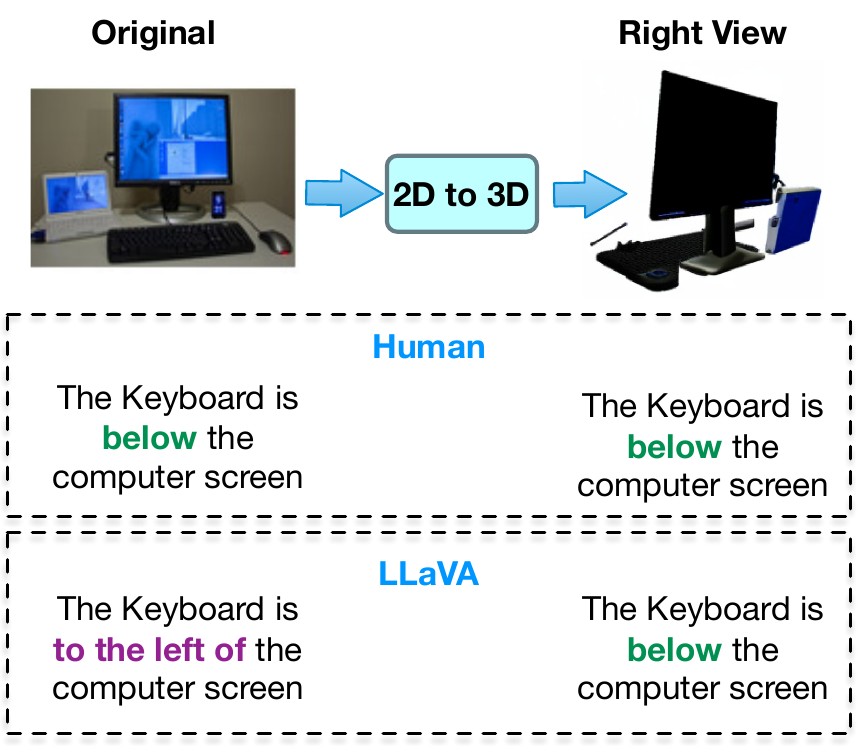}
	\caption{An example of the \vqa{} task, where humans can easily recognize positions under different views, but the vanilla \llava{}~\citep{liu2024visual} can only predict correctly from certain views. By performing 3D reconstruction to obtain different views of the image, we can improve \llava{}'s predictive accuracy.}
	\label{fig:motivation}
\end{figure}

However, many visual language tasks, such as visual question answering (\vqa{})~\citep{7780460,7298965,Zhou2017ScenePT,dai2018neural,hu2019language} and image segmentation~\citep{huang2023interformer}, require the ability to recognize spatial information from images. For instance, in visual question answering, a system must understand the spatial relationships between objects to correctly answer questions about an image~\citep{dai2018neural,hu2019language}. Visual spatial reasoning in \VLMs{} requires multi-modal understanding~\citep{Khattak_2023_CVPR}, cross-modal mapping~\citep{wang2023association}, inference of visual spatial relations~\citep{liu2023visual}, and integration of context comprehension~\citep{zhao2023survey}. This is because \VLMs{} must comprehend intricate spatial relationships within images, which are hierarchical and multifaceted, involving various relationships among multiple objects (such as containment, proximity, overlap, etc.)~\citep{chen2024spatialvlm}. Enhancing visual spatial reasoning capabilities of models can significantly elevate their performance in image comprehension and processing~\citep{yang2024improving}.

Presently, numerous studies explore various benchmarks and methods to enhance the visual spatial reasoning capabilities of \VLMs{} to improve their performance, exemplified by visual spatial reasoning~\citep{liu2023visual}, Whatsup\_vlm~\citep{kamath2023s} and SpatialVLM~\citep{chen2024spatialvlm}. These endeavors employ diverse methodologies~\citep{su2019vl,qi2020imagebert,kiela2019supervised} to augment the visual spatial reasoning abilities of models. However, despite these efforts yielding improvements in model performance to some extent, substantial challenges persist in comprehending complex scenes and intricate spatial relationships. Existing methodologies predominantly rely on 2D information, impeding the comprehensive capture of 3D spatial relationships among objects. For example, Figure \ref{fig:motivation} illustrates an example of the \vqa{} task. While humans could easily recognize positions under different views of the objects, \llava{} can only predict correctly from certain views due to the lack of image views during their pretraining.

To tackle this challenge, we propose a novel model, called \ours{}, to solve the visual spatial reasoning task through the 3D reconstruction technique, entailing the reconstruction of images in 3D and capturing them from various views. In particular, our \ours{} utilize the Zero-1-to-3~\citep{liu2023zero} model for constructing 3D views from a single 2D image within our tested datasets. This allows \VLMs{} to access richer spatial information through 3D transformation, thereby enhancing their visual spatial reasoning capabilities. We conducted comprehensive experiments across four visual spatial reasoning datasets, and the results show that all the tested \VLMs{} have notably improved spatial reasoning capabilities through our \ours{}.

Our contributions are summarized as follows:

\begin{itemize}
	\item We proposed the \ours{} model, a new model for visual spatial reasoning tasks, which can generate images of different views based on the original image to enhance spatial reasoning ability.
\end{itemize}
\begin{itemize}
	\item Our \ours{} utilises a 3D reconstruction model to obtain different views from the input image, which enhances spatial relationships at the image level to improve the visual spatial reasoning capabilities of \VLMs{}.
\end{itemize}
\begin{itemize}
	\item We validate the effectiveness of our \ours{} through visual spatial reasoning experiments conducted on four different datasets.
\end{itemize}

\begin{figure*}[t]
	\begin{minipage}[t]{0.48\linewidth}
		\centering
		\includegraphics[width=1.0\textwidth]{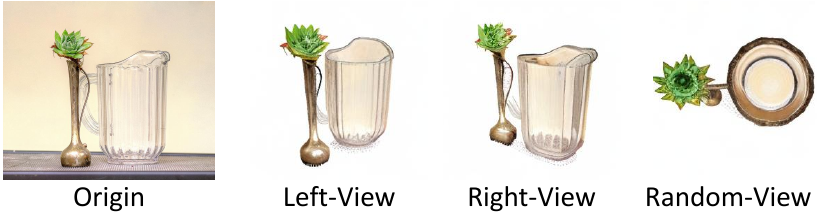}
		\subfloat{(a) Single-View}
	\end{minipage}
	\hfill
	\begin{minipage}[t]{0.48\linewidth}
		\centering
		\includegraphics[width=1.0\textwidth]{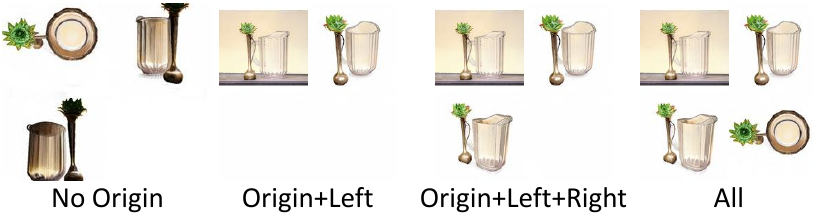}
		\subfloat{(b) Multi-View}
	\end{minipage}
	\caption{Single-view images were generated using Zero-1-to-3 to produce left-view, right-view, and random-view images. Multi-view images were created by combining these different single-view images in various configurations.}
	\label{fig:views}
\end{figure*}

\section{Related Work}

\textbf{Vision Language Models (\VLMs{}).}  \VLMs{} amalgamate computer vision and natural language processing technologies to comprehend and generate correlations between images and natural language~\citep{dosovitskiy2020image}. \VLMs{} can accept image input to generate a corresponding response in text~\citep{menon2022visual}. In terms of training architecture, recent works include joint pretraining architectures (e.g. OFA~\cite{wang2022ofa} and VLMo~\citep{bao2021vlmo}), which train image and text data jointly, and image-to-text mapping architectures (e.g., PaLI~\citep{chen2022pali}), which map an image encoder to a well-pretrained text encoder. The latter approach has gained more popularity. \VLMs{} exhibit cross-modal understanding capabilities, enabling them to extract information from images and translate it into natural language, or retrieve information from natural language and generate corresponding images~\citep{chen2022improving}. This wide range of capabilities broadens the potential of \VLMs{} (e.g., GlaMM~\citep{rasheed2023glamm} and \Mini{}~\citep{zhu2023minigpt}) in understanding and generating associations between images and language, such as Image Captioning~\citep{zhang2021vinvl}, Visual Question Answering~\citep{yang2022vision}, and Multi-modal Translation~\citep{Li2019UnicoderVLAU}. In this study, we aim to explore the visual-spatial recognition capabilities of these \VLMs{}.

\noindent\textbf{Visual Spatial Reasoning.} Visual spatial reasoning refers to the cognitive ability to understand and manipulate the spatial relationships of objects or elements in a given environment~\citep{liu2023visual}. This reasoning ability involves not only recognizing the properties of individual entities, but also understanding the complex relationships and structures between them. Recent works have aimed to benchmark the problem of visual spatial reasoning, notably \vsr{}~\citep{liu2023visual} and Whatsup\_vlm~\citep{kamath2023s}. \vsr{}~\citep{liu2023visual} evaluates the ability of \VLMs{} using text-image pairs to describe various visual spatial relationships, while Whatsup\_vlm~\citep{kamath2023s} assesses the visual spatial reasoning of \VLMs{} through specific prepositions and perspectives. Although these methods provide valuable insights into the visual spatial reasoning capabilities of \VLMs{}, they remain somewhat limited. In our work, we employ 3D reconstruction to more comprehensively test the visual spatial reasoning ability of \VLMs{}.

\noindent\textbf{3D Reconstruction.} Visual spatial reasoning encompasses the cognitive ability to comprehend and manipulate visual spatial information, while 3D reconstruction involves generating a three-dimensional representation of an object or scene from two-dimensional images or other sensor data. For instance, MonoScene~\citep{cao2022monoscene} infers dense geometric structures and semantic information of a scene from a single monocular RGB image, offering an efficient and innovative approach to complete 3D semantic scenes. VoxFormer~\citep{li2023voxformer} proposes a transformer-based~\citep{vaswani2017attention} semantic scene completion framework, addressing camera-based 3D semantic scene completion by introducing sparse voxel queries~\citep{guchait2021probing} and masked autoencoder design~\citep{he2022masked}. 
Zero-1-to-3~\citep{liu2023zero} is a 3D reconstruction model that excels in zero-shot generalization on out-of-distribution datasets and unprocessed images by using a single image as input without requiring additional 3D or depth information. We chose this model for its simplicity and effectiveness compared to MonoScene~\citep{cao2022monoscene} and VoxFormer~\citep{li2023voxformer}.

\begin{figure*}[htp]
	\centering
	\includegraphics[width=12cm]{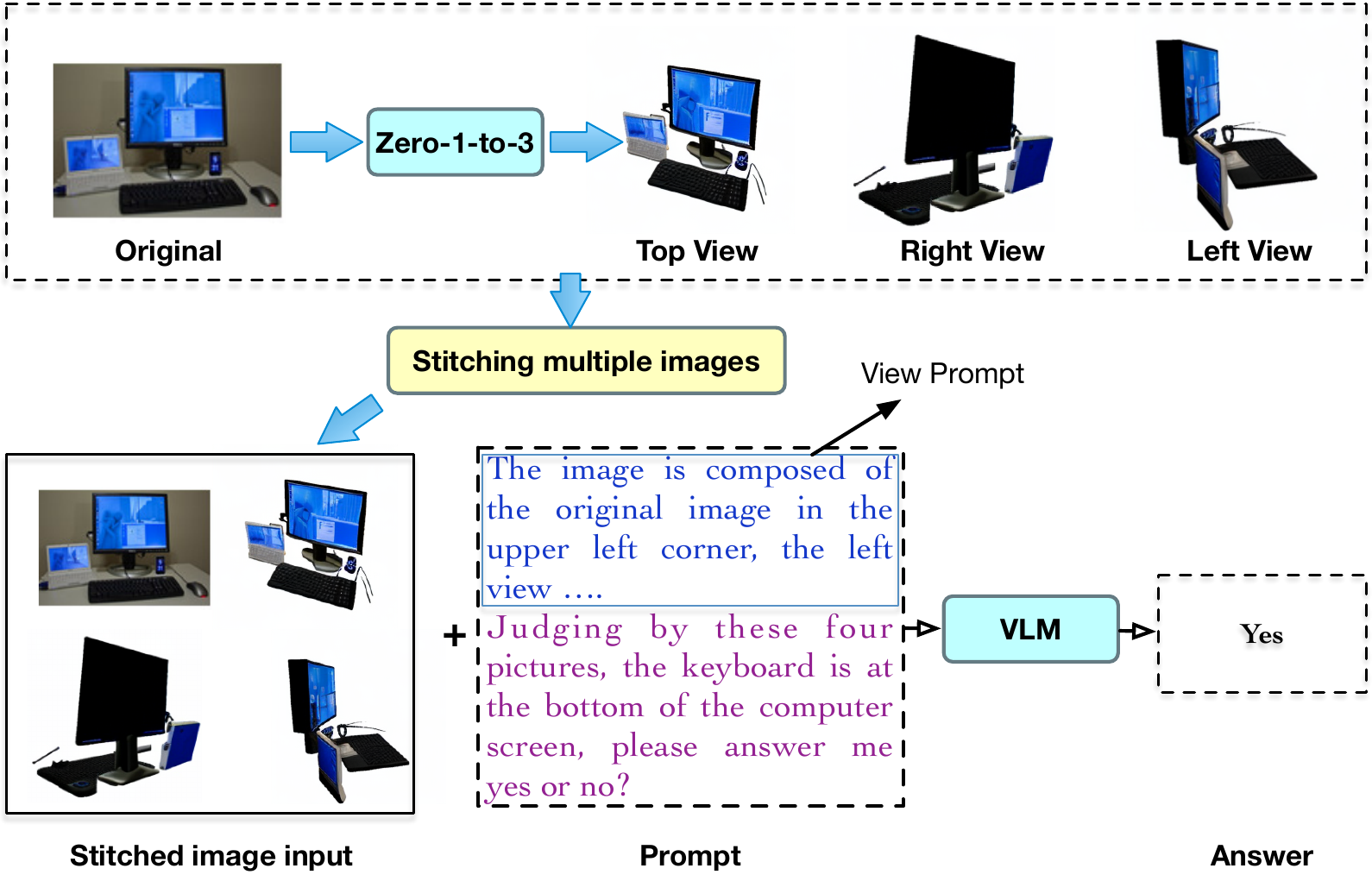}
	\caption{An overview of our proposed \ours{} model. Our \ours{} first uses Zero-1-to-3 to perform 3D reconstruction to obtain different views of the input image, and then it stitches the original images with these different views to obtain the stitched image, which is the input of a \VLM{} for answer prediction.}
	\label{fig:overview}
\end{figure*}

\section{Preliminaries}

In this section, we will formally define the task of visual spatial reasoning and introduce the Zero-1-to-3 model~\citep{liu2023zero}, which is one of the components utilized by our model.

\subsection{Visual Spatial Reasoning Task}
Visual spatial reasoning is critically important in the fields of computer vision, artificial intelligence, and machine learning~\citep{zakari2022vqa,rajabi2023towards,liu2022causal}. The visual spatial reasoning task involves inferring and understanding objects, scenes, or spatial relationships, addressing location, orientation, and spatial arrangement for accurate object localization and recognition. This task is normally formulated as a visual question answering task~\citep{liu2021visually,kamath2023s} by asking and answering questions about objects on images and through context in text. Formally, we denote an image and its associated question as \(\textit{i}\) and \(q\), respectively. The questions in this task typically ask to identify the relationships among multiple objects or to locate specified objects within the image. Figure \ref{fig:motivation} shows an example of this task. The goal of this task is to answer this question by accurately comprehending the visual spatial relations and discerning the visual spatial position of target objects relative to others.

\textbf{
	\subsection{Zero-1-to-3}
}
Zero-1-to-3~\citep{liu2023zero} is a model designed for visual reconstruction tasks. It works by identifying objects in input image \(\textit{i}\) and adjusting the camera perspective of these objects. Given an image \(i \in \mathbb{R}^{H \times W \times 3}\) and a relative camera rotation matrix \(R \in \mathbb{R}^{3 \times 3}\) and a translation vector \(T \in \mathbb{R}^{3}\), Zero-1-to-3 leans a model \(f\) such that its output is a new perspective image under the specified camera transformation:

\begin{equation}
	\hat{i}_{R,T} = f(i, R, T)
\end{equation}

where \(\hat{i}_{R, T}\) represents the generated new perspective image. When we input the original image \(\textit{i}\) and hope to generate a 3D reconstructed view from the left at an angle of 45\textdegree{} through Zero-1-to-3, we can achieve this effect by setting \(R\) to the rotation matrix of 45\textdegree{} around the Y axis and setting \(T\) to the vector translated to the left by a certain distance. In order to synthesize new views under such partial constraints, Zero-1-to-3 uses a large-scale diffusion model to integrate the geometric prior knowledge of the input image. The conditional diffusion model embedded in Zero-1-to-3 ~\citep{yang2023diffusion,podell2023sdxl} learns to control the relative camera viewpoint using synthetic datasets, which helps generate new images with specified camera transformations, such as fixed camera viewpoints (such as left or right view) or randomly generated camera viewpoints. Despite being trained on synthetic datasets, Zero-1-to-3 still demonstrates strong zero-shot generalization capabilities to out-of-distribution datasets and real-world images. Figure \ref{fig:views} shows the different views generated by Zero-1-to-3 based on the input image.

\section{Methodologies}

\subsection{Overview of \ours{}}
To address the visual spatial reasoning task, we propose \ours{}, a model that leverages Zero-1-to-3 to infer various views of input images and employs a \VLM{} to generate answers using the combined multiview images and our specially designed \textit{view prompts}. The overview of our model is illustrated in Figure \ref{fig:overview}. In particular, \ours{} is a novel visual-language model that combines large language models (such as \llama{}~\citep{touvron2023llama}) with high-level vision models, aiming to enhance visual-spatial reasoning capabilities by leveraging 3D reconstruction techniques. Given an image \(\textit{i}\), \ours{} uses Zero-1-to-3 to infer its various views \(\hat{i}_{R, T}\) and performs image stitching from multiple views, thereby providing richer spatial information. The 3D reconstruction feature provided by Zero-1-to-3 can enable \ours{} to better understand and infer spatial relationships by viewing objects from different angles. At the same time, a specialized \textit{view prompt} is designed into \ours{} to further enhance its visual spatial reasoning capabilities. This prompt helps guide the model in focusing on relevant spatial views and improving its accuracy in interpretation. The architecture of our \ours{} is shown in Figure \ref{fig:overview}, with a detailed description of its components described in the following subsections.

\subsection{Data Augmentation by 3D Reconstruction}
In our work, we use Zero-1-to-3 for 3D reconstruction on the dataset to generate \(\hat{i}_{R, T}\) from different viewpoints, e.g. the left, right and random views. Our investigation not only explores whether the creation of single-view images can enhance the visual spatial reasoning capabilities of \VLMs{} but also examines whether multi-view images can help this improvement, where multi-view images are synthesized from different \(\hat{i}_{R, T}\). Because multi-view images can provide richer spatial information, the model can get the opportunity to observe the same scene from different angles, thereby capturing a more comprehensive spatial layout and the relationship between objects, and different views can provide redundant information, so that the model can still make accurate judgments when facing noise or partial information loss. Therefore, in our work, after inferring different views of the input images, we further construct multi-view images from these single views by stitching them and testing their effectiveness against the single-view images.  Figure \ref{fig:views} shows an example of the multi-view image constructed by stitching multiple single-view images generated through Zero-1-to-3.

By synthesizing these various single-view and multi-view images, we aim to conduct comprehensive controllable experiments to determine their effectiveness in improving the spatial reasoning abilities of \VLMs{}.

\subsection{View Prompt}
\begin{figure*}[htp]
	\centering
	\includegraphics[width=12cm]{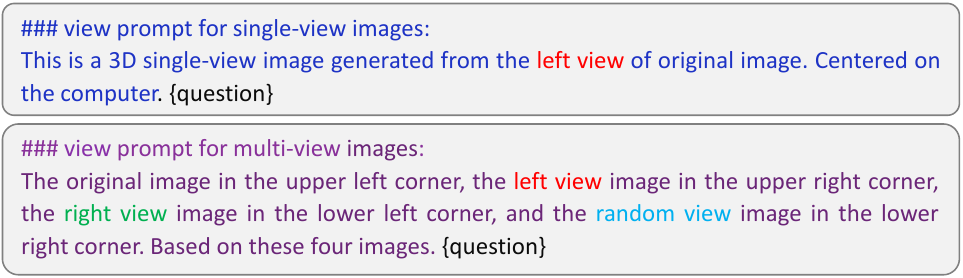}
	\caption{These view prompts are manually constructed by us. View prompt comparison between single-view images and multi-view images. \{question\} is the corresponding question in the prompt.}
	\label{fig:New_Prompt}
\end{figure*}

In our work, we first test whether the 3D reconstruction of images can improve the accuracy of \VLM{}'s visual spatial ability at the image level. To further explore whether the context prompt can enhance \VLM{}'s visual spatial reasoning ability, we introduced a special prompt called \textit{view prompt} in the experiment. This view prompt varies depending on the input image and its views. Figure \ref{fig:New_Prompt} shows two view prompt examples over a single view and multiple views of the input.

\begin{figure}[htp]
	\centering
	\includegraphics[width=8cm]{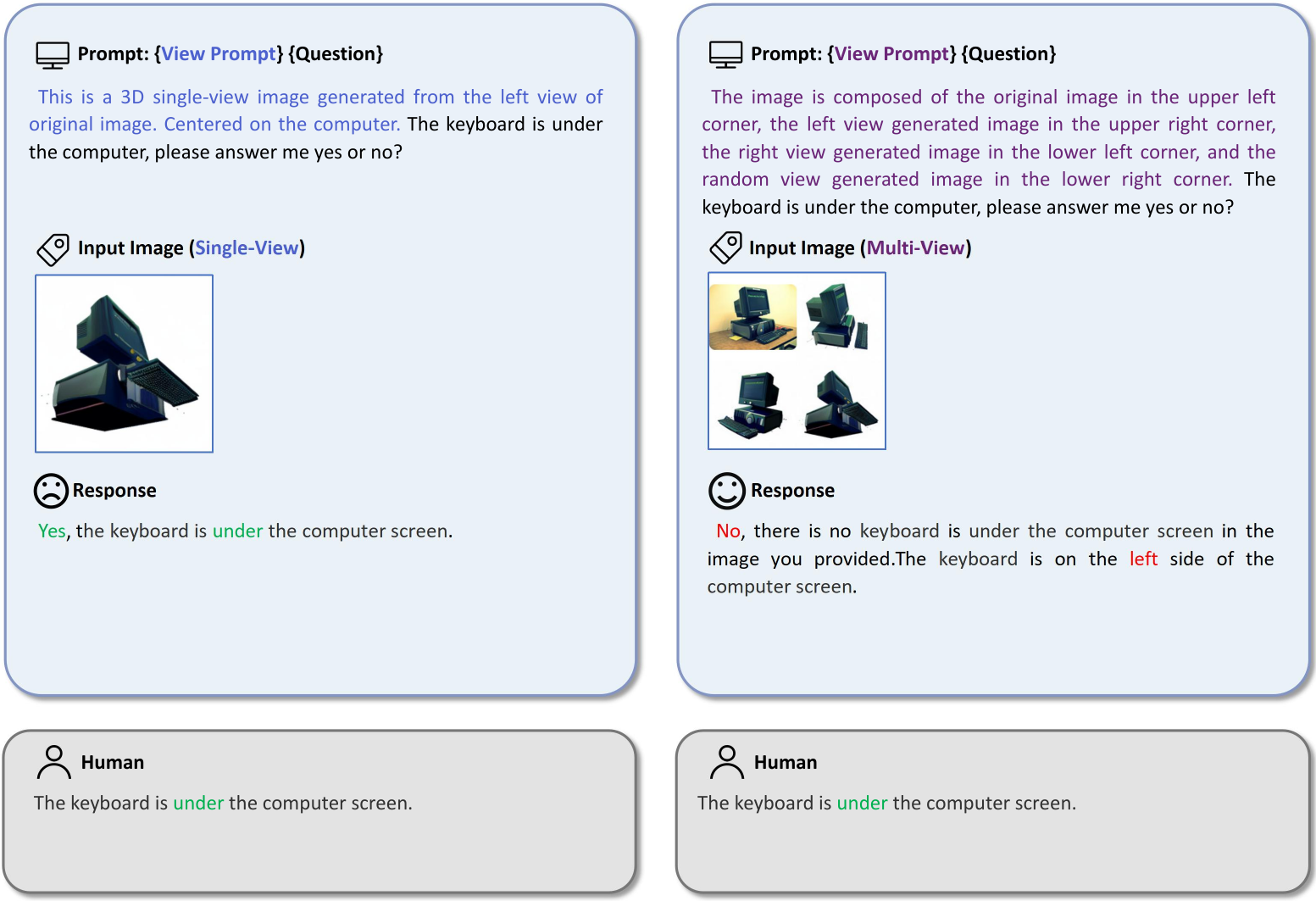}
	\caption{For both single-view and multi-view datasets, we employed a view prompt. The aim was to explore whether perspective language models could enhance their visual spatial reasoning abilities through improvements at the textual level.}
	\label{fig:prompt}
\end{figure}

We designed a variety of view prompts based on the content of different view images to guide \VLM{} to better understand and reason about the view spatial relationship between the target object and other objects in the image. We first used Zero-1-to-3~\citep{liu2023zero} to perform 3D reconstruction of \vsr{}~\citep{liu2023visual} dataset and \WU{}~\citep{kamath2023s} dataset from different viewpoints. Then during the inference of VLM, we use a prompt consisting of the question, the view prompt, and the stitched view image, to generate the answer. Figure \ref{fig:prompt} provides detailed examples of the process.

\section{Experiments}
\subsection{Experimental Setup}
\textbf{Datasets.} We selected two datasets for investigation: the \textbf{Visual Spatial Reasoning (\vsr{})}~\citep{liu2023visual} dataset and the \textbf{\WU{}} ~\citep{kamath2023s} dataset. The statistics of the two datasets are summarized in Table \ref{tab:DataSet}. The \vsr{} dataset focuses on a wide range of spatial relations and linguistic phenomena, emphasizing various visual spatial relations and describing these relations. The \WU{} dataset focuses on household items and uses different prepositions to describe spatial relations. Both datasets cover a variety of visual spatial relations (such as ``below'', ``in front of'', ``beside'', etc.).

\begin{table}[th]
	\centering
	\scalebox{1.2}{%
		\begin{tabular}{lcccc}
			\toprule
			\textbf{DataSet} & \textbf{Train} & \textbf{Development} & \textbf{Test} & \textbf{Total} \\
			\midrule
			\vsr{} (Random Split)  & 7680 & 1097 & 2195 & 10972 \\
			\vsr{} (Zero shot)     & 4713 & 231 & 616 & 5560 \\
			\WU{} (SubSet-A) & 200 & 110 & 111 & 421 \\
			\WU{} (SubSet-B) & 200 & 108 & 100 & 408 \\
			\bottomrule
		\end{tabular}
	}
	\caption{The number of images in various datasets.}
	\label{tab:DataSet}
\end{table}

\noindent\textbf{Baselines.} In our research, we employ \llava{} and \Mini{} as the primary \VLMs{}. The LLaVA version we use is LLaVA-v1.5-13B and the MiniGpt-4 version is MiniGPT-4 (Vicuna 13B). \llava{} utilizes multi-modal language-image instruction data for instruction adjustment, employing CLIP-ViT-L-336px~\citep{liu2023improved} as the visual encoder and MLP projection as the visual language cross-modal connector, achieving comprehensive understanding of both visual and language inputs. The architecture of \Mini{} involves aligning a frozen visual encoder with a frozen high-level language model, e.g. Vicuna~\citep{vicuna2023}, through projection layers, ensuring correct alignment of visual features with the high-level large language model. \Mini{} can exhibit advanced multi-modal capabilities similar to those of \GPT{}~\citep{achiam2023gpt}.

\noindent\textbf{Backbones.} We use \llava{} and  \Mini{} as the backbone \VLMs{} of our \ours{} in all the experiments. In particular, we denote \ours{} (L) as our model based on the \llava{}~\citep{liu2024visual} model, which can process image inputs and improve efficiency and performance through joint learning of image and text instructions, while  \ours{} (M) as our model based on the \Mini{}~\citep{zhu2023minigpt}) model, which combines the powerful generation ability of language models with visual information capabilities.

\noindent\textbf{Task Setting.} We classify visual spatial relationship questions for \VLMs{} into two types using our dataset split, which is applied consistently across all four datasets. The first type disrupts the visual spatial relationship between the target object and another object (for example, the correct image description is ``The apple is to the left of the banana, '' while our description is ``The apple is above (below, to the right of, in front of, etc.) the banana''). The second type describes visual spatial relationships involving objects not present in the image but related to the target object (for example, the correct image description is ``The apple is to the left of the banana, '' but our description is ``The watermelon (an object that does not exist besides the apple) is above the banana''). These classifications are employed to evaluate the visual spatial reasoning abilities of \VLMs{} and highlight the differences between our dataset and the original datasets.

\noindent\textbf{Evaluation Metric.} For evaluation, we judge the accuracy of the visual spatial reasoning ability of \ours{} (L) and \ours{} (M) based on the answers answered by \ours{} (L) and \ours{} (M). The accuracy of \ours{} (L) and \ours{} (M) responses determines their accuracy in identifying these relations. Since both datasets involve binary classification tasks, we use accuracy as the evaluation metric.

\noindent\textbf{Implementation Details.} We obtain new datasets through 3D reconstruction, consisting of single-view and multi-view datasets. In the single-view dataset, ``Origin'' refers to the original, unprocessed data, ``Left'' refers to the data processed by 3D reconstruction from the left viewpoint, and ``Right'' refers to the data processed by 3D reconstruction from the right viewpoint. ``Random'' refers to the data processed from a random viewpoint. In the multi-view dataset, there are two types of ``Multi-view'', one is multi-view without original images, and the other is multi-view with original images. Multi-view without original images is composed of single-view images, while multi-view with original images is composed of original images plus different single-view images. Our goal is to explore whether the visual spatial reasoning ability of \VLMs{} increases or decreases with the addition or omission of original images.

\begin{table*}[ht]
	\centering
	\scalebox{1.0}{%
		\begin{tabular}{lcccc}
			\toprule
			\textbf{Model} & \textbf{\vsr{} (Random Split)} & \textbf{\vsr{} (Zero Shot)} & \textbf{\WU{} (A)} & \textbf{\WU{} (B)} \\
			\midrule
			Human           & \textbf{95.40}  & \textbf{95.40}  & \textbf{100}   & \textbf{100}   \\
			CLIP (frozen)   & 56.00  & 54.50 &  58.00   &  57.50   \\
			CLIP (FT)       & 65.10 &  63.30   &  61.20   &  59.80   \\
			VisualBERT      & 55.20 & 51.00 &  53.40   &  55.60   \\
			ViLT            & 69.30 & 63.00 &  60.20   &  63.60   \\
			LXMERT          & 70.10 & 61.20 &  58.30   &  54.70   \\
			\textbf{\ours{} (M)}       & 53.80 & 52.43 & 55.76 & 53.84 \\
			\textbf{\ours{} (L)}            & \underline{70.29} & \underline{70.94} & \underline{71.74} & \underline{80.76} \\
			\bottomrule
		\end{tabular}
	}
	\caption{Accuracy performance of the compared models. Missing values indicate that \ours{} (L) and \ours{} (M) were not tested on the visual spatial reasoning on the \vsr{} and \WU{} datasets. Bests are in \textbf{bold}, and second bests are \underline{underlined}.}
	\label{tab:overall_performance}
\end{table*}

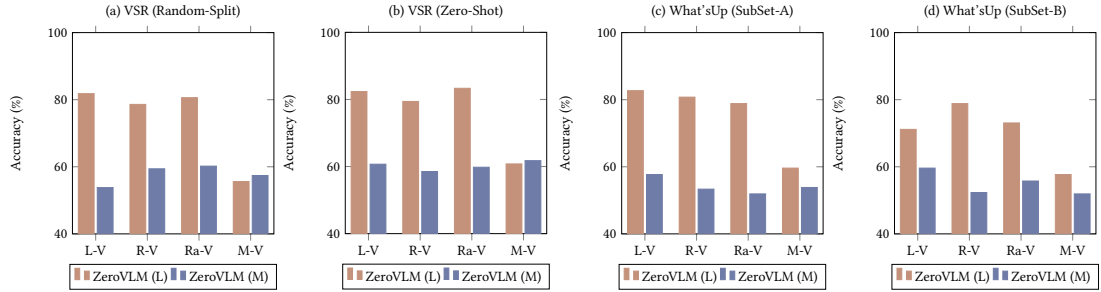
\begin{figure*}[ht]
	\centering
	\scalebox{0.6}{%
		\begin{minipage}{0.4\textwidth}
			\begin{tikzpicture}
				\begin{axis}[
					ybar,
					bar width=.35cm,
					width=\textwidth,
					height=\textwidth,
					title={(a) \vsr{} (Random-Split)},
					ylabel={Accuracy (\%)},
					symbolic x coords={L-V, R-V, Ra-V, M-V},
					xtick=data,
					ymin=40, ymax=100,
					enlarge x limits=0.15,
					legend style={at={(0.5,-0.15)}, anchor=north, legend columns=-1}
					]
					\addplot[draw=customorange,fill=customorange] coordinates {(L-V, 81.80) (R-V, 78.60) (Ra-V, 80.60) (M-V, 55.60)};
					\addplot[draw=customblue,fill=customblue] coordinates {(L-V, 53.80) (R-V, 59.40) (Ra-V, 60.20) (M-V, 57.40)};
					\legend{\ours{} (L), \ours{} (M)}
				\end{axis}
			\end{tikzpicture}
		\end{minipage}%
		\begin{minipage}{0.4\textwidth}
			\begin{tikzpicture}
				\begin{axis}[
					ybar,
					bar width=.35cm,
					width=\textwidth,
					height=\textwidth,
					title={(b) \vsr{} (Zero-Shot)},
					ylabel={Accuracy (\%)},
					symbolic x coords={L-V, R-V, Ra-V, M-V},
					xtick=data,
					ymin=40, ymax=100,
					enlarge x limits=0.15,
					legend style={at={(0.5,-0.15)}, anchor=north, legend columns=-1}
					]
					\addplot[draw=customorange,fill=customorange] coordinates {(L-V, 82.35) (R-V, 79.41) (Ra-V, 83.33) (M-V, 60.78)};
					\addplot[draw=customblue,fill=customblue] coordinates {(L-V, 60.70) (R-V, 58.52) (Ra-V, 59.80) (M-V, 61.76)};
					\legend{\ours{} (L), \ours{} (M)}
				\end{axis}
			\end{tikzpicture}
		\end{minipage}
		\centering
		\begin{minipage}{0.4\textwidth}
			\begin{tikzpicture}
				\begin{axis}[
					ybar,
					bar width=.35cm,
					width=\textwidth,
					height=\textwidth,
					title={(c) \WU{} (SubSet-A)},
					ylabel={Accuracy (\%)},
					symbolic x coords={L-V, R-V, Ra-V, M-V},
					xtick=data,
					ymin=40, ymax=100,
					enlarge x limits=0.15,
					legend style={at={(0.5,-0.15)}, anchor=north, legend columns=-1}
					]
					\addplot[draw=customorange,fill=customorange] coordinates {(L-V, 82.69) (R-V, 80.76) (Ra-V, 78.84) (M-V, 59.61)};
					\addplot[draw=customblue,fill=customblue] coordinates {(L-V, 57.69) (R-V, 53.34) (Ra-V, 51.92) (M-V, 53.84)};
					\legend{\ours{} (L), \ours{} (M)}
				\end{axis}
			\end{tikzpicture}
		\end{minipage}%
		\begin{minipage}{0.4\textwidth}
			\begin{tikzpicture}
				\begin{axis}[
					ybar,
					bar width=.35cm,
					width=\textwidth,
					height=\textwidth,
					title={(d) \WU{} (SubSet-B)},
					ylabel={Accuracy (\%)},
					symbolic x coords={L-V, R-V, Ra-V, M-V},
					xtick=data,
					ymin=40, ymax=100,
					enlarge x limits=0.15,
					legend style={at={(0.5,-0.15)}, anchor=north, legend columns=-1}
					]
					\addplot[draw=customorange,fill=customorange] coordinates {(L-V, 71.15) (R-V, 78.84) (Ra-V, 73.07) (M-V, 57.69)};
					\addplot[draw=customblue,fill=customblue] coordinates {(L-V, 59.61) (R-V, 52.33) (Ra-V, 55.76) (M-V, 51.92)};
					\legend{\ours{} (L), \ours{} (M)}
				\end{axis}
			\end{tikzpicture}
		\end{minipage}
	}
	\caption{L-V represents the left-view images, R-V represents the right-view images, Ra-V represents the random-view images, and M-V represents the multi-view images (excluding the original image). Note that here we use \ours{} (L) and \ours{} (M) for fair comparison.}
	\label{ZeroVLM_MiniGpt-4}
\end{figure*}

\subsection{Experimental Result}
\noindent\textbf{Overall performance.}  Our experimental aims to comprehensively evaluate the visual spatial reasoning capabilities of \VLMs{} \ours{} (L) and \ours{} (M) on various datasets. We seek to understand how different view datasets (including original, single-view, and multi-view datasets) affect the performance of these models in visual-spatial reasoning tasks. We first test \ours{} (L) and \ours{} (M) on the original \vsr{} dataset (without 3D reconstruction) and we reproduced LXMERT~\citep{tan2019lxmert}, VisualBERT~\citep{li2019visualbert}, and CLIP~\citep{radford2021learning} using the same parameters as in \vsr{}~\citep{liu2023visual}, and also tested them on the original \vsr{} dataset (without 3D reconstruction). The result of this comparison is reported in Table \ref{tab:overall_performance}. Our experimental results show a baseline comparison of \ours{} (L) and \ours{} (M) testing the original dataset with \vsr{}~\citep{liu2023visual}, highlighting the basic performance of \ours{} (L) and \ours{} (M).

Additionally, we tested single-view 3D reconstructions from the \vsr{} and \WU{} datasets, and multi-view datasets derived from these single views. We also assessed their visual spatial reasoning accuracy on single-view and multi-view datasets using different view prompts for \ours{}.

\noindent\textbf{Single View vs Multiple View.} In this experiment, we investigated the impact of single-view versus multi-view 3D reconstruction on the visual spatial reasoning capabilities of \VLMs{}. We assessed two \VLMs{}: \ours{} (L) and \ours{} (M), using both single-view and multi-view datasets. The results, shown in Figure \ref{ZeroVLM_MiniGpt-4}, reveal that \ours{} excels in single-view conditions with an average accuracy of 79.28\%, but its performance drops to 58.42\% in multi-view conditions. \ours{} (M) performs less well overall, with average accuracy of 56.95\% for single-view and 56.23\% for multi-view. Overall, \ours{} (L) outperforms \ours{} (M) in visual spatial reasoning tasks, particularly in single-view scenarios. While \ours{} (M) possesses unique advantages in handling multi-modalities, there is still room for improvement in specific visual spatial reasoning tasks~\citep{Battaglia2018RelationalIB}. The reason that multi-view models failed might be due to the fact that there are too many elements in the multi-view image dataset and the model cannot distinguish them correctly.

\begin{table*}[ht]
	\centering
	\scalebox{0.8}{%
		\begin{tabular}{ccccccc}
			\hline
			\textbf{View Type} & \textbf{View} & \textbf{View Prompt}  & \textbf{\vsr{} (Random Split)} & \textbf{\vsr{} (Zero Shot)} & \textbf{\WU{} (A)} & \textbf{\WU{} (B)} \\
			\hline
			& Origin &$\times$ & 70.29 & 70.94 & 71.14 & \underline{80.76} \\
			& L-V &$\times$ & 81.80 & 82.35 & 82.69 & 71.15 \\
			& R-V &$\times$ & 78.60 & 79.41 & 80.76 & 78.84 \\
			Single-View & Ra-V &$\times$ & 80.60 & 83.33 & 78.84 & 73.07 \\
			& L-V &$\checkmark$ & \textbf{89.20} & 88.63 & \underline{88.62} & 76.32 \\
			& R-V &$\checkmark$ & 87.60 & \underline{89.41} & \textbf{90.38} & \textbf{84.21} \\
			& Ra-V &$\checkmark$ & \underline{88.40} & \textbf{90.09} & 87.42 & 79.56 \\
			\hline
			& M-V &$\times$ & 55.60 & {60.78} & 59.61 & 57.69 \\
			Multi-View & Origin + L-V &$\checkmark$ & 64.60 & 50.98 & 63.46 & {65.23} \\
			& Origin + L-V + R-V &$\checkmark$ & {67.40} & {61.76} & {69.23} & {68.38} \\
			& Origin + M-V &$\checkmark$ & {72.20} & 59.80 & {65.38} & 63.21 \\
			\hline
		\end{tabular}
	}
	\caption{Performance comparison of  \ours{} (L) under different view combinations. L-V represents the left-view images, R-V represents the right-view images, Ra-V represents the random-
		view images, and M-V represents the multi-view images (excl. the original image). Bests are in \textbf{bold}, and second bests are \underline{underlined}.}
	\label{tab:compare_views}
\end{table*}

\noindent\textbf{Effect on different view prompts.} In this experiment, we aim to explore the potential enhancement of visual spatial reasoning abilities in \VLMs{} by employing different view prompts. We sought to investigate if leveraging contextual connections through textual cues could significantly improve the performance of \VLMs{} in understanding and reasoning about visual spatial relationships. To achieve this, we applied various view prompts with \ours{} and assessed its visual spatial reasoning abilities across both single-view and multi-view datasets. These view prompts were designed to provide contextual cues that could aid the model in interpreting visual spatial information more effectively. The results of this comprehensive assessment are presented in Table \ref{tab:compare_views}. The experiment demonstrated that the use of different prompts can indeed enhance the visual spatial reasoning capabilities of \VLMs{}. Overall, our study concludes that employing a view prompt can enhance the visual spatial reasoning abilities of \VLMs{}. 

\begin{figure}[h]
	\centering
	\includegraphics[width=8cm]{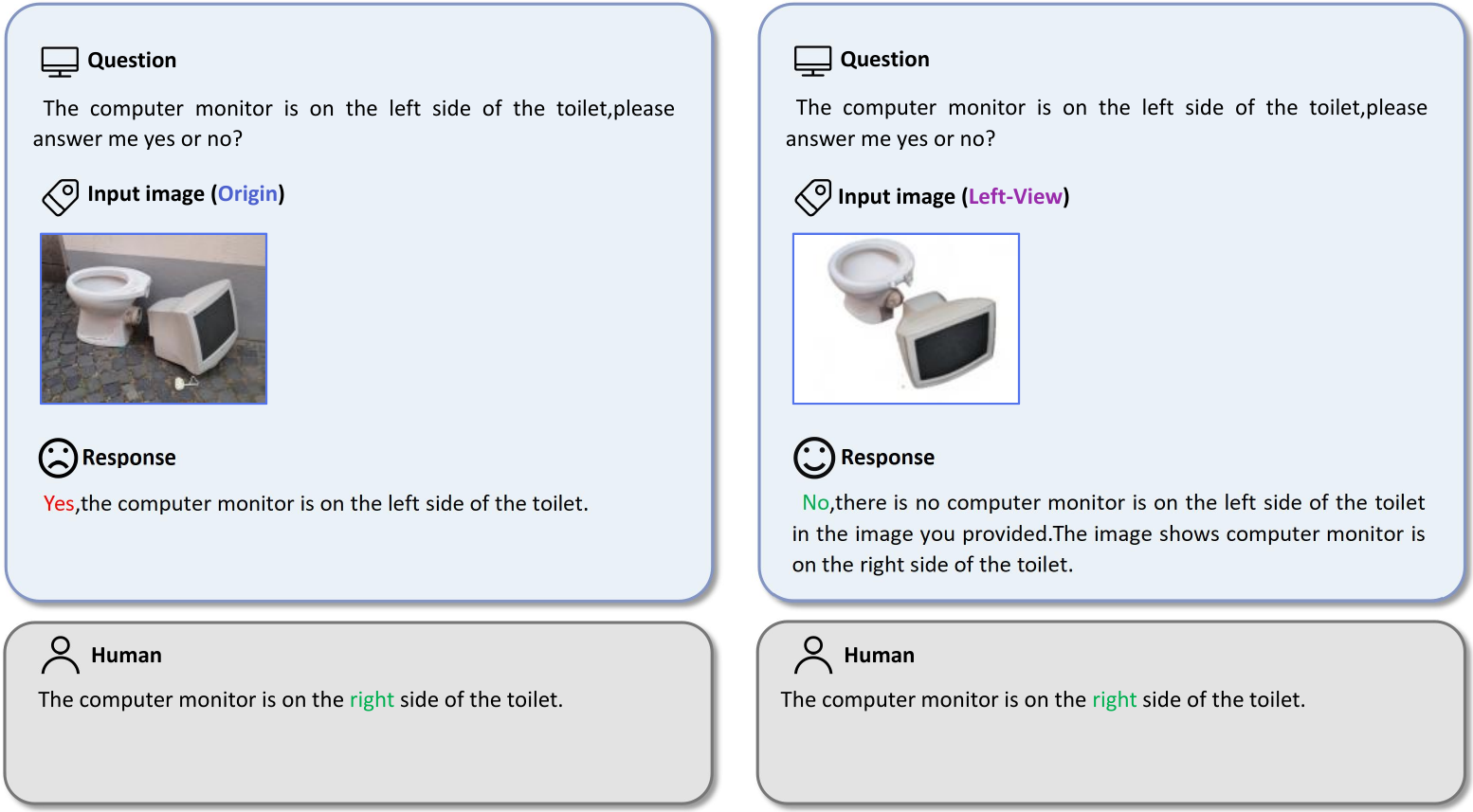}
	\caption{Different results were obtained by \ours{}
		when evaluating datasets with and without 3D reconstruction. View prompts are omitted to save space.}
	\label{fig:case_study_1}
\end{figure}

\begin{figure}[H]
	\centering
	\includegraphics[width=8cm]{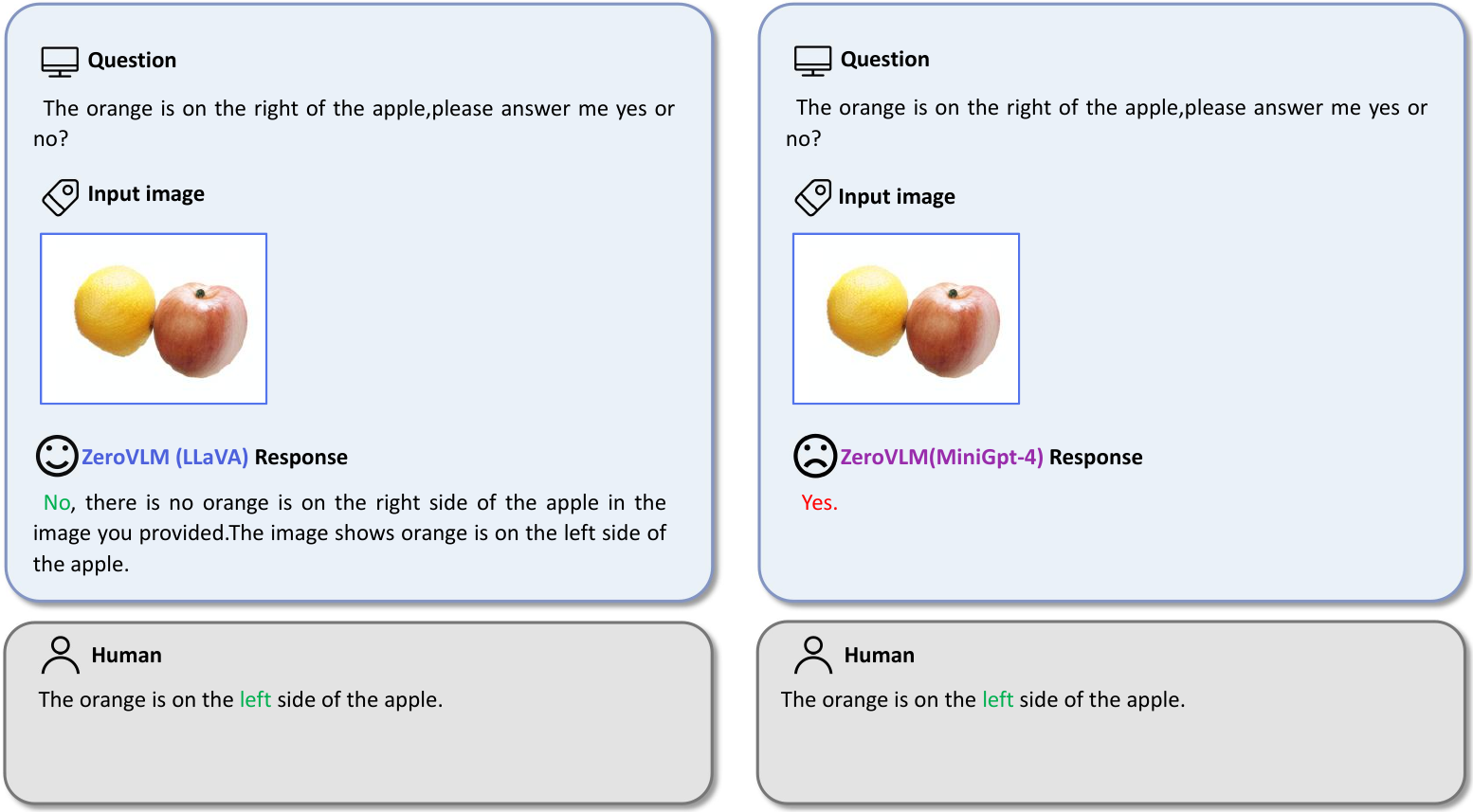}
	\caption{Different results were obtained when inputting the 3D reconstructed datasets into \ours{} (L) and \ours{} (M). View prompts are omitted to save space.}
	\label{fig:case_study_2}
\end{figure}

\noindent\textbf{Case study.} Our study aims to investigate the impact of 3D reconstruction on the visual spatial reasoning capabilities of \VLM{}. We assess whether 3D reconstruction (Zero-1-to-3) enhances these capabilities from the original dataset to the single-view dataset and compare the performance of \ours{} (L) and \ours{} (M) using single-view images after 3D reconstruction. The comparative analysis focuses on the visual spatial reasoning capabilities of \ours{} on both datasets. Figure \ref{fig:case_study_1} illustrates the enhancement by \ours{} using the single-view dataset. Figure \ref{fig:case_study_2} presents the performance comparison between \ours{} (L) and \ours{} (M). Results show that \ours{} (L) has improved visual spatial reasoning with the single-view dataset after 3D reconstruction and outperforms \ours{} (M). This suggests that 3D reconstruction provides useful spatial information aiding visual reasoning.

\section{Conclusion}
This work studies the task of visual spatial reasoning and presents a simple and novel model, called \ours{}, for enhancing the visual spatial reasoning capability of \VLMs{} through 3D reconstruction. In particular, \ours{} uses the Zero-1-to-3 model to generate different views over different angles of the input image, and uses VLMs to generate spatial reasoning answers with the stitched view image and a specially designed view prompt.  We validate the effectiveness of our \VLMs{} by comparative experiments from various views, including single-view and multiple-view images, and evaluated the performance of \llava{} and \Mini{} under our \ours{}. The experimental results suggest that 3D reconstruction with a view prompt from a single perspective can effectively enhance the model's visual spatial reasoning ability.  Future research could also develop models that dynamically adjust view prompts based on task requirements and integrate additional modal information, such as video and audio, to enhance multi-modal processing capability. 

\vfill

\section{Limitations}
Although \ours{} demonstrates significant performance improvements, we still identify the following limitations of our work: (1) Data dependency: \ours{} used specific visual spatial reasoning datasets in our experiments. Although these datasets cover a variety of scenarios, they may not fully represent all the complex spatial relationships in the real world; (2) Diversity of datasets: Although we used multiple datasets for testing, these datasets may not fully cover all possible application scenarios. Therefore, the model may need further training and adjustment to ensure its generality and robustness when processing different types of images and tasks; (3) The generalization ability of the model: Although the Zero-1-to-3 model performs well in zero-shot generalization, in some extreme cases, the model may still not accurately capture the spatial relationship.  Therefore, considering other methods for the 3D reconstruction module is one of our future work directions.

\section{Potential risks}
The development and deployment of \VLMs{} for visual spatial reasoning present several potential risks. One significant risk is the dependency on specific datasets for training, which may not encompass the full diversity of spatial relationships encountered in real-world scenarios. This can lead to models that perform well in controlled environments but fail to generalize effectively. Additionally, the computational resources required for 3D reconstruction and multi-view generation are substantial, posing scalability and real-time application challenges. Another concern is the potential for bias in the datasets, which can result in models that unfairly represent or perform poorly on underrepresented spatial arrangements or object types. Furthermore, privacy and security issues arise when handling large amounts of visual data, necessitating strict adherence to data protection guidelines. Lastly, there are ethical considerations regarding the misuse of enhanced spatial reasoning capabilities for surveillance or other invasive applications, underscoring the need for transparency and accountability in the deployment of these technologies. Addressing these risks is essential to ensure the effective and responsible use of \VLMs{} in visual spatial reasoning.

\bibliography{sample-base}
\bibliographystyle{ACM-Reference-Format}

\appendix

\section{Appendix}
\begin{figure}[H]
	\centering
	\includegraphics[width=7.5cm]{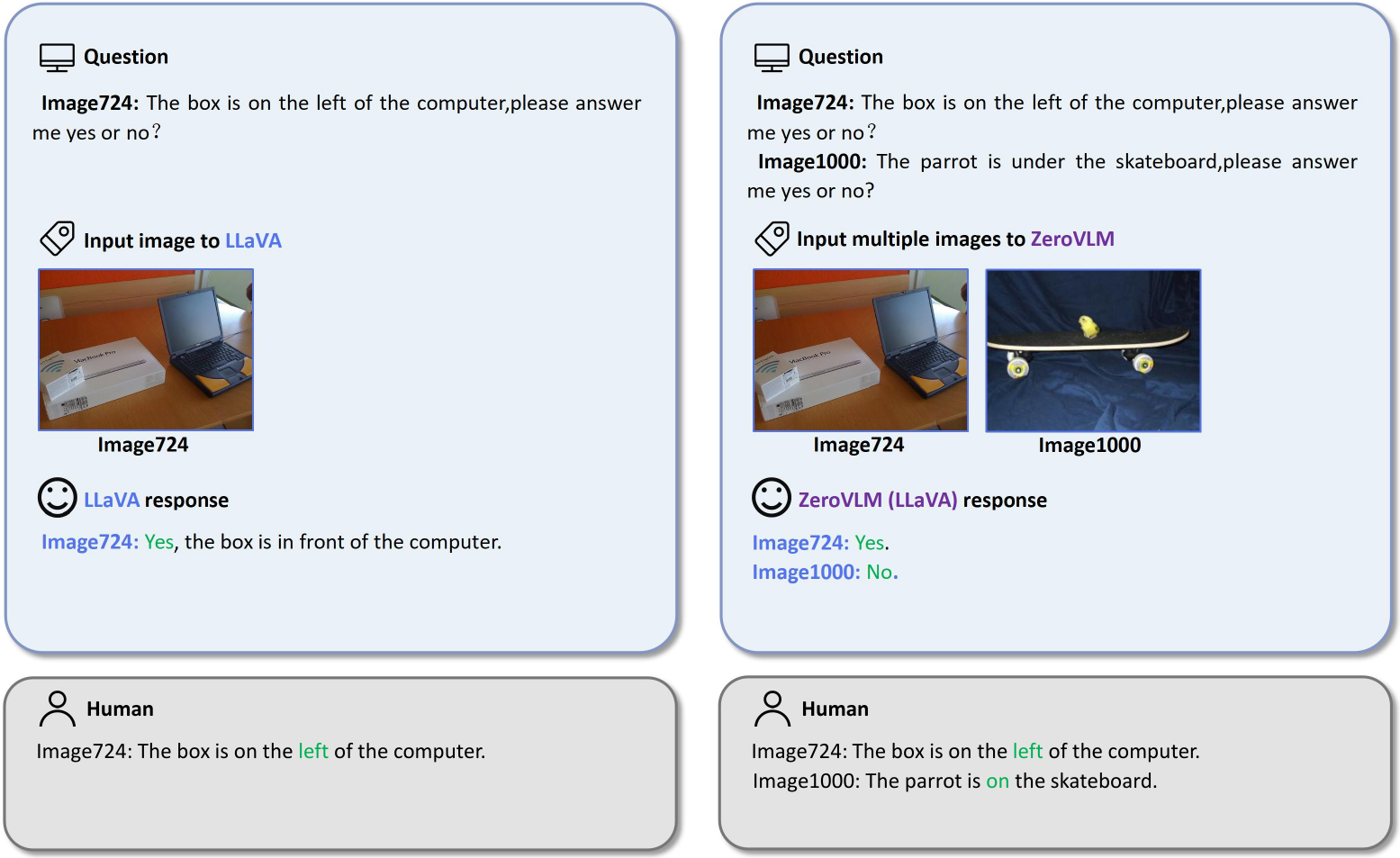}
	\caption{The primary difference between \llava{} and \ours{} (L) lies in their input handling. \llava{} processes a single image as input, whereas \ours{} (L) is designed to handle multiple images. Due to the limited expressive capacity of individual images, we restrict the input for \ours{} (L) to two images.}
	\label{fig:LLaVA_ZeroVLM}
\end{figure}

\begin{figure}[h]
	\centering
	\includegraphics[width=6.5cm]{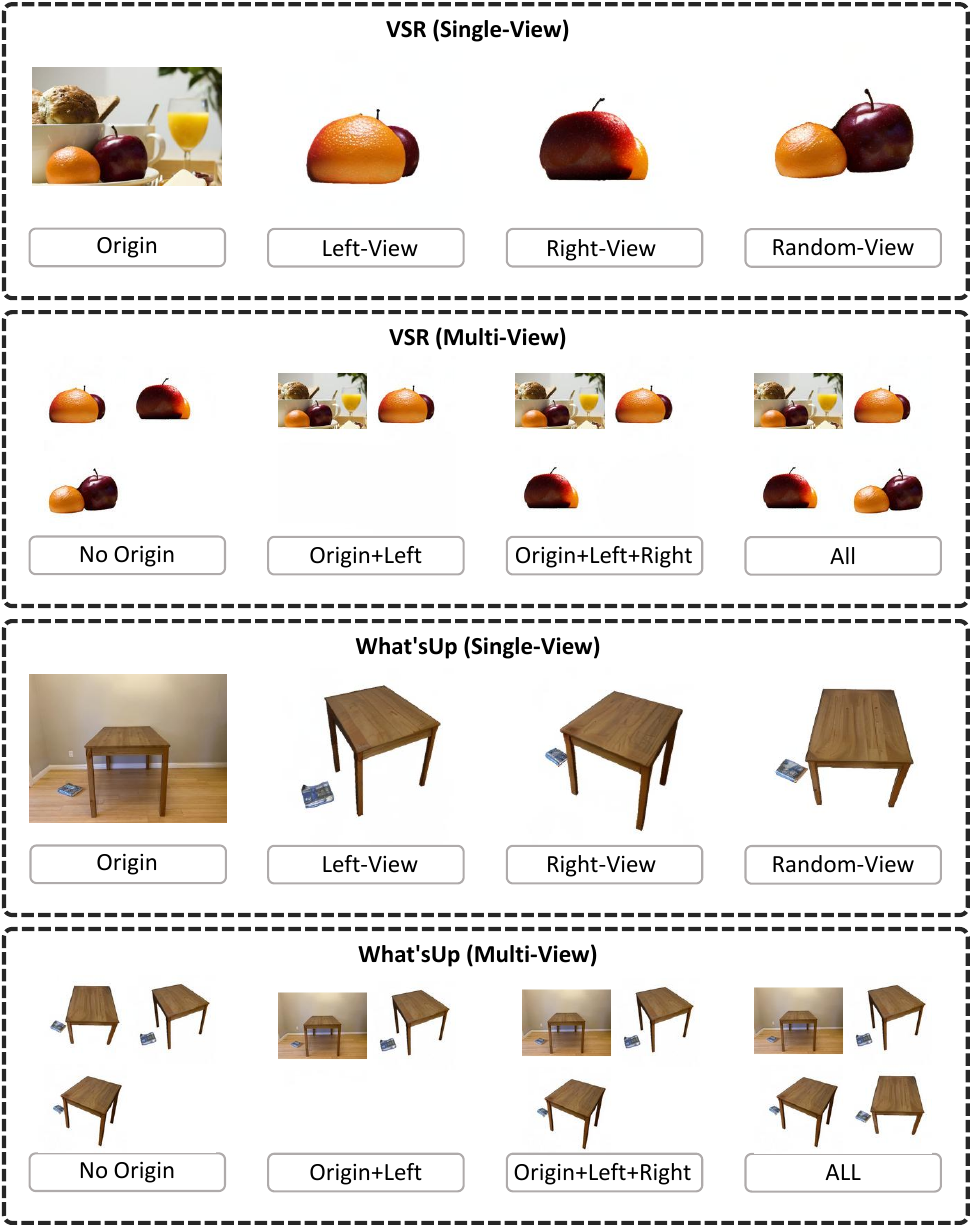}
	\caption{The 3D reconstructed single-view and multi-view versions of each dataset are presented.}
	\label{fig:data_stat}
\end{figure}

This section contains additional results. Figure \ref{fig:LLaVA_ZeroVLM} illustrates the differences between \llava{} and \ours{} (L). In our study, we prepare the required questions in \json{} file format and modify \ours{} (L) to recognize and accept \json{} file inputs. We enable \ours{} (L) to identify the corresponding questions by matching the input image names with the image names in the \json{} file. Figure \ref{fig:data_stat} illustrates the single-view and multi-view datasets obtained after 3D reconstruction of the \vsr{} dataset and the \WU{} dataset.

\end{document}